\documentclass{article}
\usepackage{arxiv}
\usepackage{graphicx,epsfig}
\usepackage{dcolumn}
\usepackage{xcolor}
\usepackage{lineno}
\usepackage{amsfonts}
\usepackage{amssymb}
\usepackage{mathtools}
\usepackage{mathabx}
\usepackage{bm}
\usepackage{qcircuit}
\usepackage{braket}
\usepackage{float}

\usepackage{pdfpages}
\usepackage{times}

\usepackage{soul}
\usepackage{url}
\usepackage{booktabs}
\usepackage{amsmath}
\usepackage{amsthm}
\usepackage[normalem]{ulem}
\usepackage[utf8]{inputenc}
\usepackage{diagbox}
\usepackage{lineno}
\usepackage{amsfonts}
\usepackage{amssymb}
\usepackage{svg}
\usepackage{graphicx}
\usepackage{mathtools}
\usepackage{mathabx}
\usepackage{dcolumn}
\usepackage{bm}
\usepackage{xcolor}
\usepackage{qcircuit}
\usepackage{braket}
\usepackage{float}
\usepackage{algorithm}
\usepackage{algorithmicx}
\usepackage{algpseudocode}

\newcommand{\norm}[1]{\left\lVert#1\right\rVert}

\newcommand{\bea}{\begin{equation}}
\newcommand{\eea}{\end{equation}\noi}
\newcommand{\ber}{\begin{eqnarray}}
\newcommand{\eer}{\end{eqnarray}\noi}
\title{Financial Vision Based Reinforcement Learning Trading Strategy}

\author{{\hspace{1mm}Yun-Cheng Tsai} \\
	Department of Data Science\\
	Soochow University\\
	\texttt{pecutsai@gm.scu.edu.tw}\\
	\And
	{\hspace{1mm}Fu-Min Szu}\\
	Department of Electrical Engineering\\
	National Taiwan University\\
	\AND
	{\hspace{1mm}Jun-Hao Chen}\\
	Department of Computer Science and\\
	Information Engineering\\
	National Taiwan University\\	
	\And
	{\hspace{1mm}Samuel Yen-Chi Chen}\\
	Computational Science Initiative\\
	Brookhaven National Laboratory
}

\begin{document}
\maketitle

\begin{abstract}
Recent advances in artificial intelligence (AI) for quantitative trading have led to its general superhuman performance in significant trading performance. However, the potential risk of AI trading is a ``black box'' decision. Some AI computing mechanisms are complex and challenging to understand. If we use AI without proper supervision, AI may lead to wrong choices and make huge losses. Hence, we need to ask about the AI ``black box'', including why did AI decide to do this or not? Why can people trust AI or not? How can people fix their mistakes? These problems also highlight the challenges that AI technology can explain in the trading field.
\end{abstract}

 
\section{\label{sec:intro}Introduction}
Suppose investors want to directly predict the future transaction price or ups and downs. In that case, the fatal assumption is that the training data set is consistent with the data distribution that has not occurred in the future. However, the natural world will not let us know whether the subsequent data distribution will change. Because of this, even if researchers add a moving window to the training process, it is inevitable that ``machine learning obstacles-prediction delay'' will occur. Our method can avoid ``machine learning obstacles-prediction delay'', We also propose auto trading by deep reinforcement learning. Our new article has the following contributions:
\begin{enumerate}
    \item Our first contribution is not to make future predictions but to focus on the current ``candlesticks pattern detection'', such as Engulfing Pattern, Morning Star,\ldots.
    \item Our second contribution focuses on detecting trading entry and exit signals combined with related investment strategies.
    \item Our third contribution found from experiments that the 15-minute price data of Ethereum train through transfer learning is suitable for US stock trading.
\end{enumerate}

With the rise of deep learning and reinforcement learning technology, breakthrough innovations are in computer trading software~\cite{zhang2020deep, xiong2018practical}. Artificial intelligence (A.I.) is more efficient than a calculation model that only uses static data~\cite{goldfarb2018ai}. Investment companies have used computer algorithms to process transactions for several years. Both new and old investment companies have begun to use artificial intelligence to help customers process investments~\cite{davenport2018artificial}.

Recent advances in A.I. for quantitative trading have led to its widespread demonstration of superhuman performance in significant trading performance~\cite{liu2020adaptive}. Likewise, reinforcement learning (R.L.)~\cite{sutton2018reinforcement} has found tremendous success and demonstrated a superhuman-level of capabilities in the trading field~\cite{deng2016deep}.

A.I. trading starts with supervised learning. The machine can judge the desired result based on the given characteristics through data characteristics and specific tags. However, in the era of big data, many data are often not so complete. Many people cannot distinguish between new A.I. transactions and traditional programming transactions. It is impossible to organize each data into a specific format, so using supervised learning-related algorithms for various situational prediction analyses is impossible. Instead, A.I. trading must base on unsupervised learning algorithms, such as deep learning and reinforcement learning. Deep learning has broken through the limitations of previous calculations due to advances in computer science and A.I. algorithms. Through deep learning and reinforcement learning, hundreds of high-level neural networks that simulate the human brain calculate.

Some A.I. computing mechanisms are complex and challenging to understand. However, the potential risk of A.I. trading is a ``black box'' decision (see~\cite{rudin2019stop, adadi2018peeking}). If we use A.I. without proper supervision, A.I. may lead to wrong choices and make huge losses. Hence, we need to ask about the A.I. ``black box'', including why did A.I. decide to do this or not? Why can people trust A.I. or not? How can people fix their mistakes? These problems also highlight the challenges that A.I. technology can explain in the trading field.

Humans can judge the candlestick pattern by instinct and determine the trading strategies with the corresponding candlestick pattern in the financial market. Could a computer think as humans see? Our goal is to teach machines to understand what they see like humans. For example, the device recognizes candlesticks' patterns, inferring their geometry, and understanding the market's relationships, actions, and intentions. In addition, we would like to know why the decisions are what A.I. models are. One must seek explanations of the process behind a model's development, not just explanations of the model itself.

Our study designs an innovative explainable A.I. trading framework by combing financial vision with deep reinforcement learning. Hence, we propose a financial vision field, which can understand the critical components of a candle, and what they indicate, to apply candlestick chart analysis to a trading strategy. We combine deep reinforcement learning to realize intuitive trading based on a financial vision to surveillance candlestick. We involve observing a large number of the candlestick, forming automatic responses to various pattern recognition.

With these extraordinary capabilities in automatic control, it is natural to consider R.L. techniques in algorithmic trading. Indeed, several works have tried to apply R.L. to trade financial assets automatically.

One of the challenges to constructing an effective RL-based algorithmic trading system is to properly encode the input signal for the agent to make decisions.
With the recent advances in convolutional neural networks (CNN), a potential scheme encodes financial time series into images.

This work proposes an algorithmic trading framework based on deep reinforcement learning and the G.A.F. encoding method. Our contributions are the following:
\begin{itemize}
    \item Provide an algorithmic trading framework for the study of RL-based strategies.
    \item Demonstrate successful R.L. trading agents with G.A.F. encoded inputs of price data and technical indicators.
\end{itemize}

The difference between A.I. trading is that A.I. is a target environment (such as the Standard \& Poor's 500). A.I. trading uses unsupervised machine competition to learn. Machine intelligence determines when to place an order or stop selling at a profit. The breakthrough of machine learning in A.I. trading is to use a new type of unsupervised learning to formulate strategies through data identification features. For example, the golden cross is a multi-feature and allows A.I. to backtest and learn through more than 400 million transaction records in 20 years. As a result, A.I. robots can find high-profit making money models and gain instant advantages through high-frequency calculations.

The global FinTech industry is already the next economic driving force for investment in many countries. Robo-advisors are already common in advanced European and American countries. We believe ordinary investors will gradually accept that financial advisors no longer rely only on high-end franchised services. Wealthy members generally serve the investing public with financial needs. The development of AI-to-AI transactions will make the financial industry a brighter future in upgrading the international A.I. transaction industry. The sector will fully upgrade to bring users a new investment and financial management experience, creating unprecedented synergies.

Reinforcement learning can interact with the environment and is suitable for applications in decision control systems. Therefore, we used the reinforcement learning method to establish a trading strategy for cryptocurrency and U.S.A. stock markets, avoiding the longstanding unstable trends in profound learning predictions. We found from experiments that the 15-minute price data of Ethereum train through transfer learning. After learning the candlesticks pattern suitable for entering and exiting the market of U.S. stock trading. Compared to the top ten most popular ETFs, the experimental results demonstrate superior performance. This study focuses on financial vision, Explainable methods, and links to their programming implementations. We hope that our paper will serve as a reference for superhuman performances and why the decisions are in the trading system.

The paper organizes as follows. In Section~\ref{sec:CandlesticksEncoding}, we introduce the concepts of financial vision for candlesticks pattern recognition. In Section~\ref{sec:Reinforcement Learning} we introduce the R.L. background knowledge used in this work. In Section~\ref{sec:GAF_RL}, we describe the proposed GAF-RL trading framework. In Section~\ref{sec:ExpAndResults}, we describe the experimental procedures and results in detail. Finally we discuss the results in Section~\ref{sec:Discussion} and conclude in Section~\ref{sec:Conclusion}.

\section{\label{sec:CandlesticksEncoding}Financial Vision}
Technical analysis is a specific product through historical data to make trading decisions or signal a general designation. The data includes price, volume, and price to get the assets on the market reaction. We try to find out the direction of the future changes. This idea is from the people's behavior in the market, which has reproducibility.

The trading account's psychological decision-making is from a large proportion of investors. By researching the past, others' trading behavior, and believing that this behavior may appear again based on experience, make a rational choice.

\subsection{Candlesticks Pattern}
Following the chart is drawn from historical prices according to specific rules. These features help traders to see the price trend. The three more common types of charts are histograms, line charts, and the most widely used candlestick.

\begin{figure}[ht]
\centering
\includegraphics[scale=0.4]{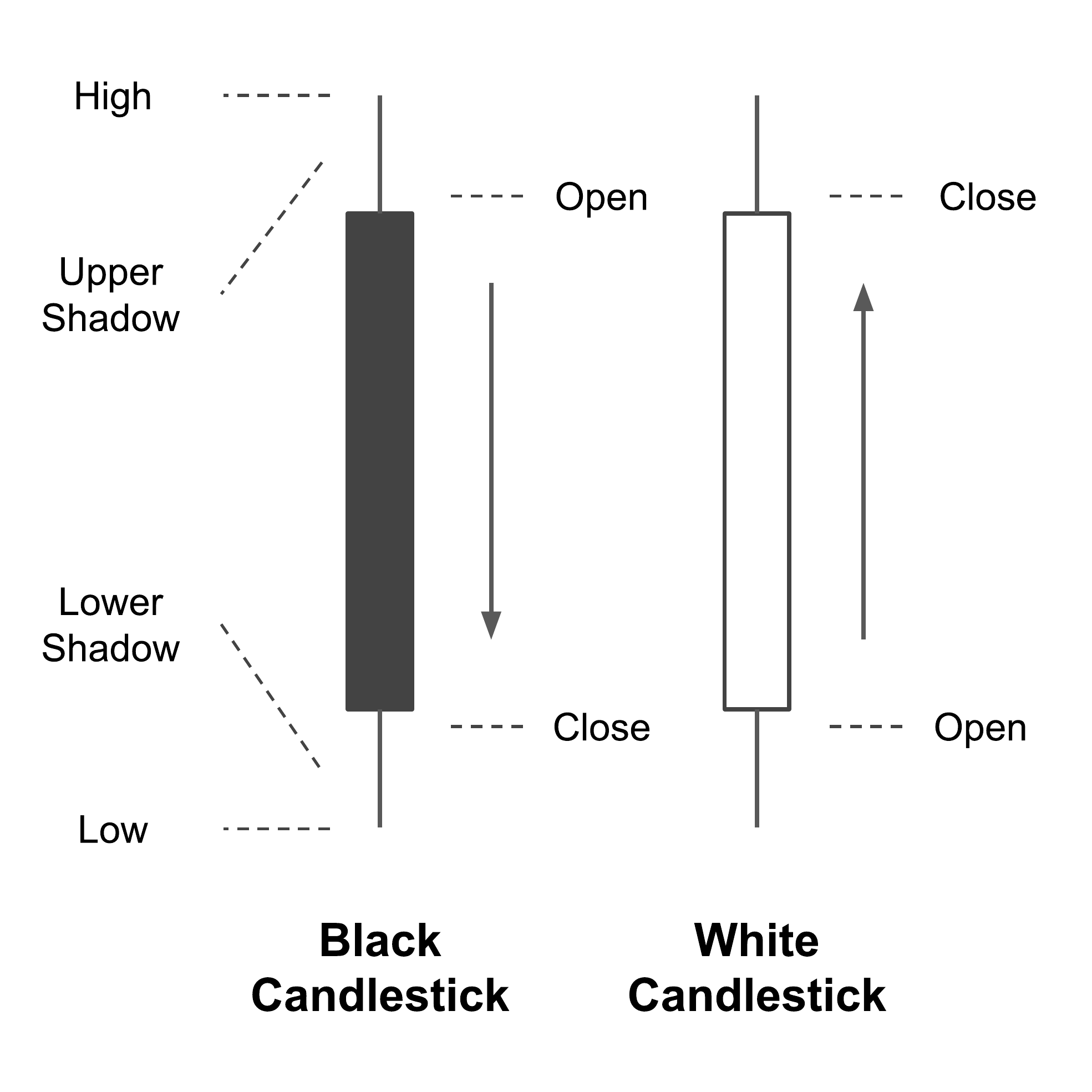}
\caption{The shape of a candlestick.}
\label{candlestick_concept}
\end{figure}

The candlestick originated from Japan in the 17th century. It has been popular in Europe and the United States for more than a century, especially in the foreign exchange market. As the most popular chart in technical analysis, traders should understand it.
It is named after a candle, as shown in Figure~\ref{candlestick_concept}.

Each bar of candlestick draws from open price, high price, low price, and close price as follows:
\begin{enumerate}
\item Open price: This price is the first price that occurs during the period;
\item High price: the highest price that occurs during the period;
\item Low price: the lowest price that occurs during the period;
\item Close price: The last price that occurs during the period.
\end{enumerate}
If the close price is higher than the open price, the candlestick follows as:
\begin{enumerate}
\item the top of the candle body is the close price;
\item the bottom is the open price;
\item the color is usually green or white.
\end{enumerate}
If the close price is lower than the open price, the candlestick follows as:
\begin{enumerate}
\item the open price above the candle body;
\item the close price below;
\item the color is usually red or black.
\end{enumerate}

The candlestick has no hatching in some cases because the open or close price coincides with the high or low price. For example, if the candle is very short, the open and close prices of the candlestick are very similar.

Focuses on the relationship between price and volume, money is the fundamental element to push prices. Even though the enormous volumes of goods do not necessarily lead to immediate price changes. However, the trading amount reflects the commodity on the market by the degree of attention, can also effectively neutralize dilute price volatility caused by the artificial manipulation, belong to the steady and lower the risk of loss of judgment strategy, usually accompanied by
other strategy target reference.

\subsection{Gramian Angular Field (GAF) Encoding}
In this work, we employ the Gramian Angular Field (GAF)~\cite{wang2015imaging} method to encode the time series into images. 
Firstly, the time-series $X=\{{x}_{1}, {x}_{2} \cdots {x}_{n}\}$ to be encoded is scaled into the interval $[0, 1]$ via the minimum-maximum scaling in Equation~\ref{equ:minmax}.

\begin{align}
\widetilde{x}_{i}&=\frac{x_i-\min(X)}{\max(X)-\min(X)}
\label{equ:minmax}
\end{align}
The notation $\widetilde{x}_i$ represents each normalized element from the entire normalized set $\widetilde{X}$. The arc cosine values $\phi_{i}$ of each $\widetilde{x}_i$ are calculated.
\begin{equation}
\begin{aligned}
\phi_{i} &= \arccos(\widetilde{x}_i), 0\leq \widetilde{x}_i \leq 1, \widetilde{x}_i \in \widetilde{X}
\label{equ:gaf_arccos}
\end{aligned}
\end{equation}
These $\phi$ angles use to generate the GAF matrix as follows:
\begin{equation}
\begin{aligned}
\textup{GAF} &=\cos(\phi_i + \phi_j) \\ &=
\left [\begin{matrix}
\cos(\phi_{1}+\phi_{1}) & \cdots & \cos(\phi_{1}+\phi_{n}) \\
\cos(\phi_{2}+\phi_{1}) & \cdots & \cos(\phi_{2}+\phi_{n}) \\
\vdots & \ddots & \vdots \\
\cos(\phi_{n}+\phi_{1}) & \cdots & \cos(\phi_{1}+\phi_{n})
\end{matrix}\right]
\label{equ:gaf}
\end{aligned}
\end{equation}

The generated GAF $n \times n$ matrix, where $n$ is the length of sequence considered, is used as inputs to the CNN.
This method makes it possible to keep the temporal information while avoiding recurrent neural networks, which are computationally intensive. GAF encoding methods employed in various financial time-series problems~\cite{wang2015imaging,chen2020encoding, chen2020explainable}.

\section{\label{sec:Reinforcement Learning}Reinforcement Learning}
Reinforcement learning (RL) is a machine learning paradigm in which an \emph{agent} learns how to make decisions via interacting with the environments ~\cite{sutton2018reinforcement}. 
The reinforcement learning model comprises an agent. The agent performs an action based on the current state. The action receives from the environment and returns feedback to the agent. The feedback can be either a reward or a penalty. Once the agent gets the reward, they adjust the relative function between the state and the action to maximize the overall expected return. The function could be a value function or a policy function.

A value function refers to the reward obtained from a particular action in a specific state. Therefore, accurately estimating the value function is an essential component of the model. Conversely, underestimating or overestimating the value of certain conditions or actions would influence learning performance.
A policy function is ideal for achieving a maximum expected return in a particular state. In the reinforcement learning model, actions that maximize expected return (value) in a specific condition are called policies. In several advanced models, policy functions directly apply to maximize expected returns. 

Concretely speaking, the \emph{agent} interacts with an \emph{environment} $\mathcal{E}$ over a number of discrete time steps. At each time step $t$, the agent receives a \emph{state} or \emph{observation} $s_t$ from the environment $\mathcal{E}$ and then chooses an \emph{action} $a_t$ from a set of possible actions $\mathcal{A}$ according to its \emph{policy} $\pi$. The policy $\pi$ is a function which maps the state or observation $s_t$ to action $a_t$. In general, the policy can be stochastic, meaning that given a state $s$, the action output can be a probability distribution $\pi(a_t|s_t)$ conditioned on $s_t$. After executing the action $a_t$, the agent receives the state of the next time step $s_{t+1}$ and a scalar \emph{reward} $r_t$. The process continues until the agent reaches the terminal state or a pre-defined stopping criteria (e.g. the maximum steps allowed). An \emph{episode} is defined as an agent starting from a randomly selected initial state and following the aforementioned process all the way through the terminal state or reaching a stopping criteria.

We define the total discounted return from time step $t$ as $R_t = \sum_{t'=t}^{T} \gamma^{t'-t} r_{t'}$, where $\gamma$ is the discount factor that lies in $(0,1]$. In principle, $\gamma$ is from the investigator to control how future rewards weigh the decision-making function. When we use a large $\gamma$, the agent weighs the future reward more heavily. On the other hand, future rewards are ignored quickly with a small $\gamma$. The immediate rewards will weigh more. 
The goal of the agent is to maximize the expected return from each state $s_t$ in the training process. The \emph{action-value function} or \emph{Q-value function} $Q^\pi (s,a) = \mathbb{E}[R_t|s_t = s, a]$ is the expected return for selecting an action $a$ in state $s$ based on policy $\pi$. The optimal action value function $Q^*(s,a) = \max_{\pi} Q^\pi(s,a)$ gives a maximal action-value across all possible policies. The value of state $s$ under policy $\pi$, $V^\pi(s) = \mathbb{E}\left[R_t|s_t = s\right]$, is the agent's expected return by following policy $\pi$ from the state $s$. Various RL algorithms are designed to find the policy which can maximize the value function. The RL algorithms which maximize the value function are called \emph{value-based} RL.  

\subsection{Policy Gradient}

In contrast to the \emph{value-based} RL, which learns the value function and use it as the reference to generate the decision on each time-step, there is another kind of RL method called \emph{policy gradient}. In this method, the policy function $\pi(a|s;\theta)$ is parameterized with the  parameters $\theta$. The $\theta$ will then be subject to the optimization procedure which is \emph{gradient ascent} on the expected total return $\mathbb{E}[R_{t}]$. One of the classic examples of policy gradient algorithm is the REINFORCE algorithm~\cite{williams1992simple}. In the standard REINFORCE algorithm, the parameters $\theta$ are updated along the direction $\nabla_{\theta} \log \pi\left(a_{t} | s_{t} ; \theta\right) R_{t}$, which is the unbiased estimate of $\nabla_{\theta} \mathbb{E}\left[R_{t}\right]$. However, the policy gradient method suffers from large variance of the $\nabla_{\theta} \mathbb{E}\left[R_{t}\right]$, making the training very hard. To reduce the variance of this estimate and keep it unbiased, one can subtract a learned function of the state $b_{t}(s_{t})$, which is known as the \emph{baseline}, from the return. The result is therefore $\nabla_{\theta} \log \pi\left(a_{t} | s_{t} ; \theta\right)\left(R_{t}-b_{t}\left(s_{t}\right)\right)$.
\subsection{Advantage Actor-Critic(A2C)}

A learned estimate of the value function is a common choice for the baseline $b_{t}(s_{t}) \approx V^{\pi}(s_{t})$. This choice usually leads to a much lower variance estimate of the policy gradient. When one uses the approximate value function as the baseline, the quantity $R_{t} - b_{t} = Q(s_{t}, a_{t}) - V(s_{t})$ can be seen as the \emph{advantage} $A(s_{t}, a_{t})$ of the action $a_{t}$ at the state $s_{t}$. Intuitively, one can see this advantage as ``how good or bad the action $a_{t}$ compared to the average value at this state $V(s_{t})$.'' For example, if the $Q(s_{t}, a_{t})$ equals to $10$ at a given time-step $t$, it is not clear whether $a_{t}$ is a good action or not. However, if we also know that the $V(s_{t})$ equals to, say $2$ here, then we can imply that $a_{t}$ may not be bad. Conversely, if the $V(s_{t})$ equals to $15$, then the advantage is $10 - 15 = -5$, meaning that the $Q$ value for this action $a_{t}$ is well below the average $V(s_{t})$ and therefore that action is not good. This approach is called \emph{advantage actor-critic}(A2C) method where the policy $\pi$ is the actor and the baseline which is the value function $V$ is the critic~\cite{sutton2018reinforcement}.
\subsection{Proximal Policy Optimization (PPO)}

In the policy gradients method, we optimize the policy according to the \emph{policy loss}
\[
L_{\text{policy}}(\theta) = \mathbb{E}_{t}[-\log\pi\left(a_{t}\mid s{t};\theta \right)]
\]
via gradient descent. However, the training itself may suffer from instabilities. If the step size of the policy update is too small, the training process would be too slow. On the other hand, if the step size is too large, there will be high variance in training. The proximal policy optimization (PPO) \cite{schulman2017proximal} fixes this problem by limiting the policy update step size at each training step. The PPO introduces the loss function called \emph{clipped surrogate loss function} that will constraint the policy change a small range with the help of a clip. Consider the ratio between the probability of action $a_{t}$ under current policy and the probability under the previous policy
\[
q_{t}(\theta)=\frac{\pi\left(a_{t} \mid s_{t} ; \theta \right)}{\pi\left(a_{t} \mid s_{t} ; \theta_{\text{old}}\right)}.
\]
If $q_{t}(\theta) > 1$, it means the action $a_{t}$ is with higher probability in the current policy than in the old one. And if $0 < q_{t}(\theta) < 1$, it means that the action $a_{t}$ is less probable in the current policy than in the old one. Our new loss function can then be defined as
\[
L_{\text{policy}}(\theta) = \mathbb{E}_{t}[q_{t}(\theta)  A_{t}] = \mathbb{E}_{t}[\frac{\pi\left(a_{t} \mid s_{t} ; \theta \right)}{\pi\left(a_{t} \mid s_{t} ; \theta_{\text{old}}\right)}  A_{t}],
\]
where
\[
A_{t} = R_{t} - V(s_{t}|\theta)
\]
is the advantage function. If the action under the current policy is much more probable than in the previous approach, the ratio $q_{t}$ may be significant—leading to a considerable policy update step. The original PPO algorithm \cite{schulman2017proximal} circumvents this problem by adding a constraint on the ratio, which can only be in the range $0.8$ to $1.2$. The modified loss function is now.
\[
L_{\text{policy}}(\theta) = \mathbb{E}_{t}[-min(q_{t}A_{t}, clip(q_{t}, 1 - C, 1 + C)A_{t})],
\]
where the $C$ is the clip hyperparameter (common choice is $0.2$). Finally, the value loss and entropy bonus add into the total loss function as usual:
\[
L(\theta) = L_{\text{policy}} + c_1 L_{\text{value}} - c_2 H,
\]
where
\[
L_{\text{value}} = \mathbb{E}_{t}[\|R_{t} - V(s_{t}|\theta)\|^2]
\]
is the value loss and $H = \mathbb{E}_{t}[H_{t}] = \mathbb{E}_{t}[-\sum_{j}\pi\left(a_{j} \mid s_{t} ; \theta\right) \log(\pi\left(a_{j} \mid s_{t} ; \theta\right))]$ is the entropy bonus which is to encourage exploration.

\subsection{PPO Algorithms}
\begin{center}
\scalebox{0.8}
{
\begin{minipage}{\linewidth}

\begin{algorithm}[H]
\begin{algorithmic}
\State Define the number of total episode $M$
\State Define the maximum steps in a single episode $S$
\State Define the update timestep $U$
\State Define the update epoch number $K$
\State Define the epsilon clip $C$
\State Initialize trajectory buffer $\mathcal{T}$
\State Initialize timestep counter $t$
\State Initialize two sets of model parameters $\theta$ and $\theta_{\text{old}}$
\For{episode $=1,2,\ldots,M$} 
    \State Reset the testing environment and initialise state $s_1$
    \For{step $=1,2,\ldots,S$}
        \State Update the timestep $t = t + 1$
        
        \State Select the action $a_t$ from the policy $\pi\left(a_{t} \mid s_{t} ; \theta_{\text{old}}\right)$
        
        \State Execute action $a_t$ in emulator and observe reward $r_t$ and next state $s_{t+1}$
        
        \State Record the transition  $\left(s_t,a_t,\log \pi\left(a_{t} \mid s_{t} ; \theta_{\text{old}}\right) ,r_t\right)$ in $\mathcal{T}$
        
        \If{$t = U$}
            \State Calculate the discounted rewards $R_{t}$ for each state $s_t$ in the trajectory buffer $\mathcal{T}$
            
            \For{$k = 1,2,\ldots,K$}
                \State Calculate the log probability $\log \pi\left(a_{t} \mid s_{t} ; \theta \right)$, state values $V(s_{t},\theta)$ and entropy $H_{t}$.
                
                \State Calculate the ratio $ q_{t} = \exp{(\log \pi\left(a_{t} \mid s_{t} ; \theta \right) - \log \pi\left(a_{t} \mid s_{t} ; \theta_{\text{old}} \right))}$
                
                \State Calculate the advantage $A_{t} = R_{t} - V(s_{t},\theta) $
                
                \State Calculate the $surr_1 = q_{t} \times A_{t}$ 
                
                \State Calculate the $surr_2 = clip(q_{t}, 1 - C, 1 + C) \times A_{t}$
                
                \State Calculate the loss $L = \mathbb{E}_{t}[-min(surr_1,surr_2) + 0.5 \norm{V(s_t,\theta) - R_{t}}^{2} - 0.01 H_{t}]$
                
                \State Update the agent policy parameters $\theta$ with gradient descent on the loss $L$
                
            \EndFor
            
            \State Update the $\theta_{\text{old}}$ to $\theta$
            \State Reset the trajectory buffer $\mathcal{T}$
            \State Reset the timestep counter $t = 0$
        \EndIf

    \EndFor
    
\EndFor
\end{algorithmic}
\caption{PPO algorithmic}
\label{ppo_org}
\end{algorithm}
\end{minipage}
}
\end{center}

\section{\label{sec:GAF_RL}Financial Vision Based RL Trading Framework}
\begin{figure}[ht]
\centering
\includegraphics[scale=0.35]{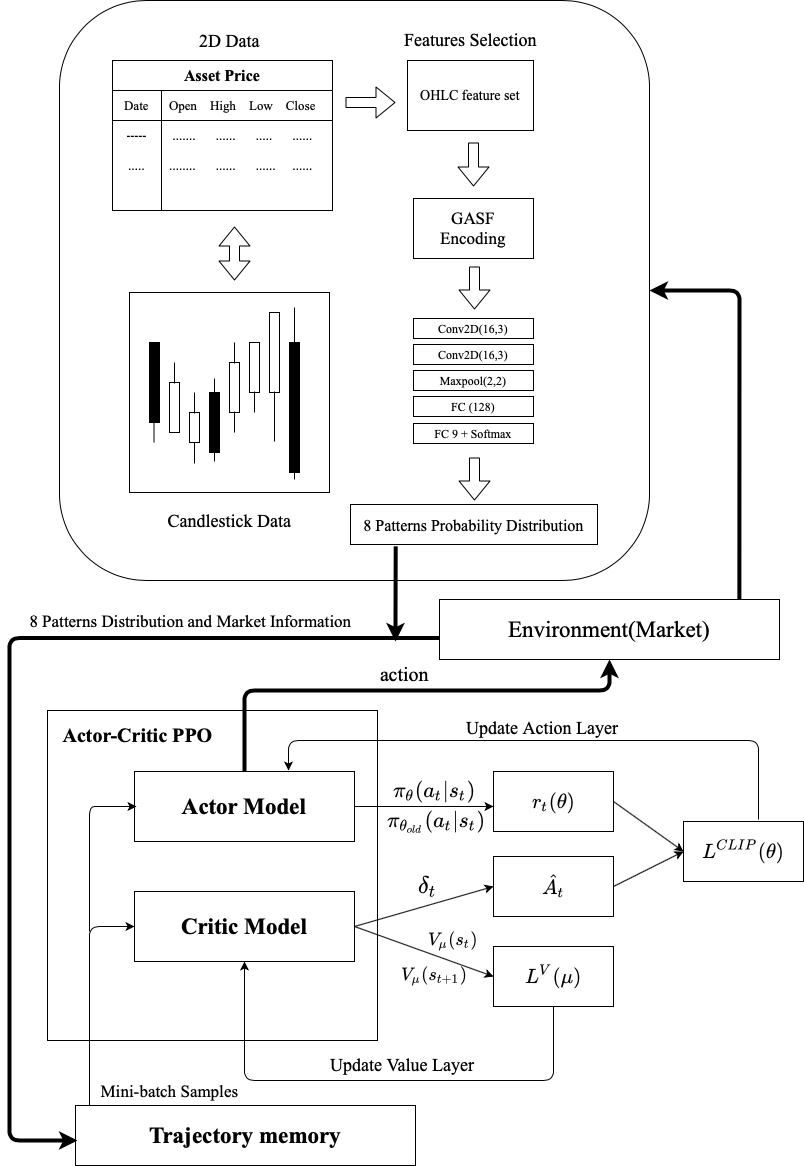}
\caption{GAF-RL Framework.}
\label{GAF-RL Framework}
\end{figure}

As the figure~\ref{GAF-RL Framework} shows, there are two large architectures, one of which is part of GAF Encoding and Actor-Critic PPO architecture. GAF Encoding aims to analyze time sequences in a way that is different from traditional time sequences. To extract the time sequence characteristics from another perspective, we must independently train pattern recognition before the experiment to make its accuracy rise to a certain level. For example, take the probability distribution of patterns and be the input model of reinforcement learning. This study used the ETH/USD exchange data in units of 15 mins from January 1, 2020, to July 1, 2020, as a reference for the system's environment design and performance calculation. The period is the training period for the agent. The agent repeatedly applied the data to learn in this period, eventually obtaining an optimal investment strategy.

However, when it comes time to train the PPO's trading strategy, initial market information and Pattens' identification information are brought into the PPO's strategy model. The market information includes technical indicators, bid strategies, and the state of open interest and volume. The PPO model takes place after a sufficient number of trajectories collect. It is input into the PPO model of the actor-critic. There are two different layers, one is the action layer, and the other is the value layer. The output layer of the action layer will have three trading signals, namely buy, sell and hold. This strategy sets a limit of the chips to three times at most. The value layer is to determine the potential value of each state and the difference with the reward $\delta_{t}$ to create the advantage function $\hat{A_t}$, $V_{\mu}(S_t)$ and $V_{\mu}(S_ {t+1})$ are simultaneously generated to update the value layer gradient.

In the  PPO training framework, off-policy training is the old policy  $\pi_{\theta_{old}}(a_{t}|s_t)$ interact with the environment produce an important sample to produce another policy gradient to update policy network $\pi_{\theta}(a_{t}|s_t)$. After sampling several times, the old policy weights to be updated as the weight of the new policy. The clipped surrogate loss function will constrain the policy change in a small range. 

To avoid too much divergence between the distribution of the two policies during training, resulting in an unstable training situation. The actions that are output by the action layer uses to interact with the following environment to establish the whole training cycle.

\subsection{Financial Vision Based PPO Algorithms}

\begin{center}
\scalebox{0.77}
{
\begin{minipage}{\linewidth}

\begin{algorithm}[H]
\begin{algorithmic}
\State Define the OHLC of asset with trade date $A$
\State Define the window sizes of trade date $W$
\State Define the total length of trade date $T_{total}$
\State Define the function GAF encoding $GAF(x)$
\State Define the function CNN model $CNN(x)$
\State Define the PPO algorithm $PPO$
\State Define the max timestep $T_{max}=T_{total}-W$

\For{$t = 1,2,\ldots,T_{max}$}
    \State Calculate the asset OHLC to GAF encoding  $A_{gaf}=GAF(A[t:t+W])$
    \State Calculate the output layer from CNN and softmax layer   $CNN(A_{gaf})$
    \State Define the probability distribution of eight patterns  $P_t=CNN(A_{gaf})$
    \State Train the $PPO$ model by environment state $P_t$
\EndFor

\end{algorithmic}
\caption{Pattern recognition scheme based on GAF and CNN }
\label{ppo_quantum_gates}
\end{algorithm}
\end{minipage}
}
\end{center}

\begin{center}
\scalebox{0.77}
{
\begin{minipage}{\linewidth}

\begin{algorithm}[H]
\begin{algorithmic}
\State Define the state of probability distribution and max timestep from pattern recognition $P,T_{max}$
\State Define the environment state with the probability distribution $s=P$
\State Define the number of total episode $M$
\State Define the maximum steps in a single episode with max timestep $S=T_{max}$
\State Define the update timestep $U$
\State Define the update epoch number $K$
\State Define the epsilon clip $C$
\State Initialize trajectory buffer $\mathcal{T}$
\State Initialize timestep counter $t$
\State Initialize two sets of model parameters $\theta$ and $\theta_{\text{old}}$
\For{episode $=1,2,\ldots,M$} 
    \State Reset the testing environment and initialize state $s_1=P_1$
    \For{step $=1,2,\ldots,S$}
        \State Update the timestep $t = t + 1$
        
        \State Select the action $a_t$ from the policy $\pi\left(a_{t} \mid s_{t} ; \theta_{\text{old}}\right)$
        
        \State Execute action $a_t$ in emulator and observe reward $r_t$ and next state $s_{t+1}=P_{t+1}$
        
        \State Record the transition  $\left(s_t,a_t,\log \pi\left(a_{t} \mid s_{t} ; \theta_{\text{old}}\right) ,r_t\right)$ in $\mathcal{T}$
        
        \If{$t = U$}
            \State Calculate the discounted rewards $R_{t}$ for each state $s_t$ in the trajectory buffer $\mathcal{T}$
            
            \For{$k = 1,2,\ldots,K$}
                \State Calculate the log probability $\log \pi\left(a_{t} \mid s_{t} ; \theta \right)$, state values $V(s_{t},\theta)$ and entropy $H_{t}$.
                
                \State Calculate the ratio $ q_{t} = \exp{(\log \pi\left(a_{t} \mid s_{t} ; \theta \right) - \log \pi\left(a_{t} \mid s_{t} ; \theta_{\text{old}} \right))}$
                
                \State Calculate the advantage $A_{t} = R_{t} - V(s_{t},\theta) $
                
                \State Calculate the $surr_1 = q_{t} \times A_{t}$ 
                
                \State Calculate the $surr_2 = clip(q_{t}, 1 - C, 1 + C) \times A_{t}$
                
                \State Calculate the loss $L = \mathbb{E}_{t}[-min(surr_1,surr_2) + 0.5 \norm{V(s_t,\theta) - R_{t}}^{2} - 0.01 H_{t}]$
                
                \State Update the agent policy parameters $\theta$ with gradient descent on the loss $L$
                
            \EndFor
            
            \State Update the $\theta_{\text{old}}$ to $\theta$
            \State Reset the trajectory buffer $\mathcal{T}$
            \State Reset the timestep counter $t = 0$
        \EndIf

    \EndFor
    
\EndFor
\end{algorithmic}
\caption{PPO for GAF-based algorithmic trading}
\label{GAF-ppo_quantum_gates}
\end{algorithm}
\end{minipage}
}
\end{center}

\section{\label{sec:ExpAndResults}Experiments and Results}
\subsection{Experiment Data}
The experiment results are from two parts as follows:
\begin{enumerate}
    \item The first part is the training patterns of GAF-CNN and
    \item the second part is the strategies of PPO-RL.
\end{enumerate}

The training patterns are from eight common candlestick patterns. Then the good enough pattern recognition is trained in advance for the second part environment of the PPO training framework.

\begin{figure}[ht]
\centering
\includegraphics[scale=0.5]{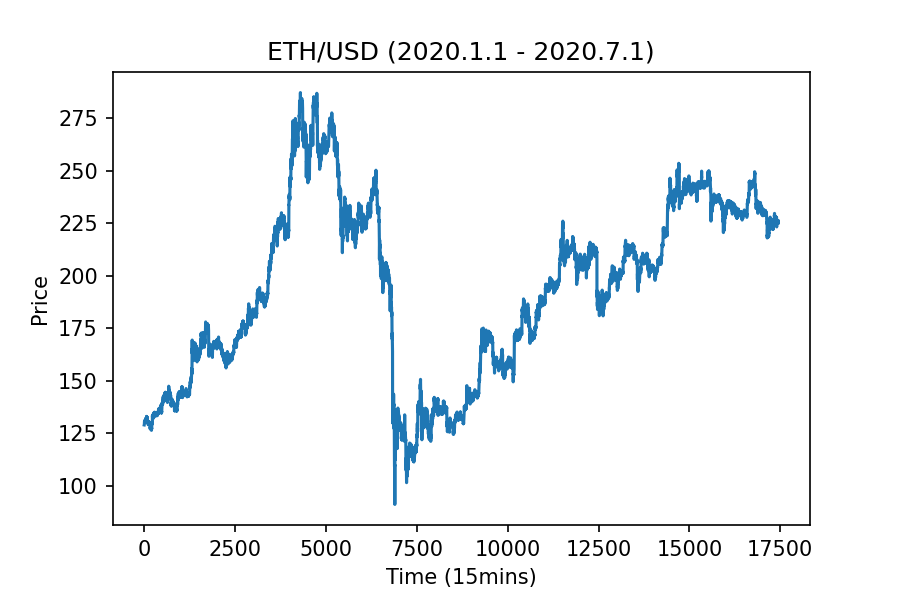}
\caption{ETH/USD 15min (2020.1.1-2020.7.1).}
\label{GAF-RL_results_ETH/USD}
\end{figure}

Figure~\ref{GAF-RL_results_ETH/USD} shows the training materials from the cryptocurrency ETH/USD 15 mins from January 1, 2020, to July 1, 2020. Due to the characteristics of this kind of commodity, it has the advantages of high volatility, continuous trading time, and significant trading volume. Therefore, there is enough information to learn more effective commodity trading strategies. 
\begin{figure}[ht]
\centering
\includegraphics[scale=0.39]{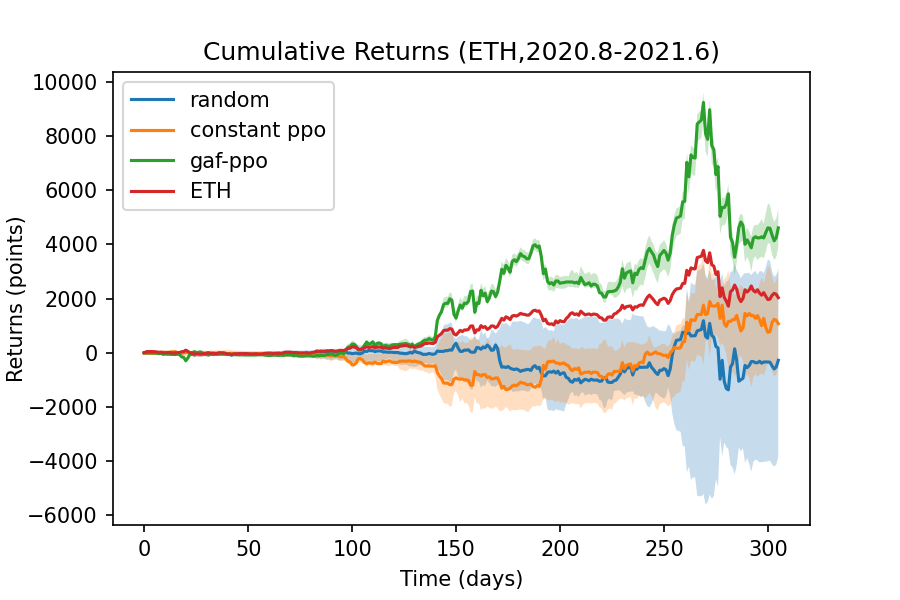}
\includegraphics[scale=0.39]{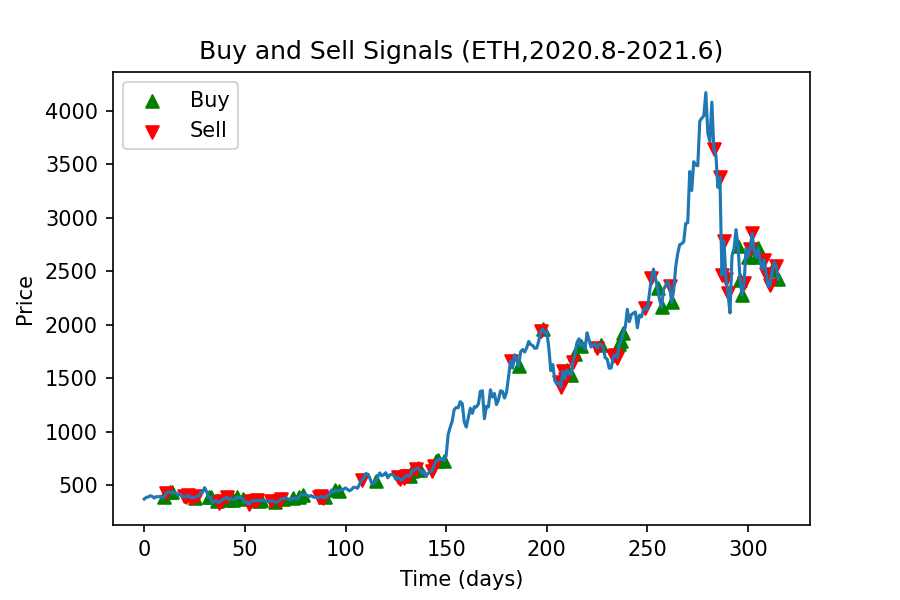}
\caption{GAF-PPO model is applied to the cryptocurrency market.}
\label{GAF-RL_results_cryptocurrency}
\end{figure}
\begin{figure}[ht]
\centering
\includegraphics[scale=0.39]{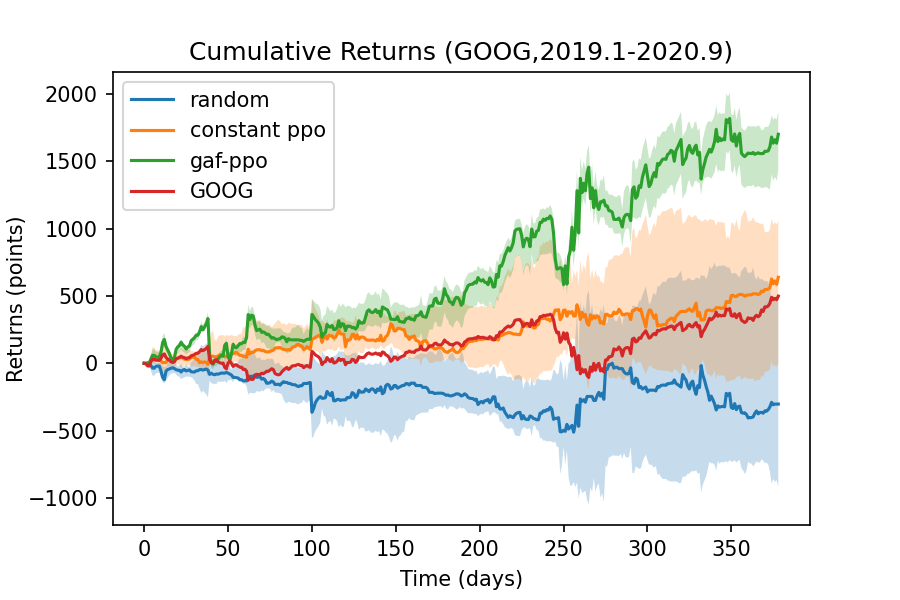}
\includegraphics[scale=0.39]{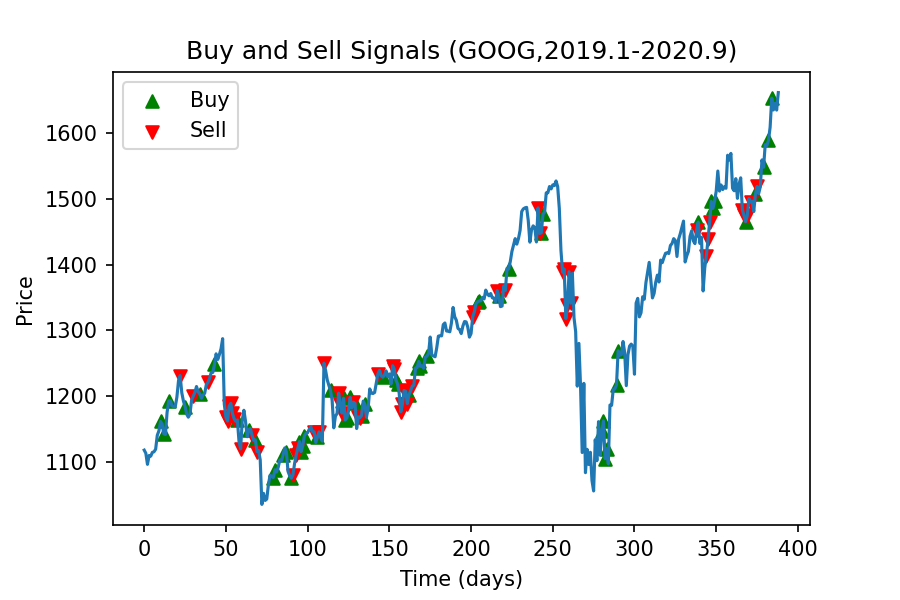}
\includegraphics[scale=0.39]{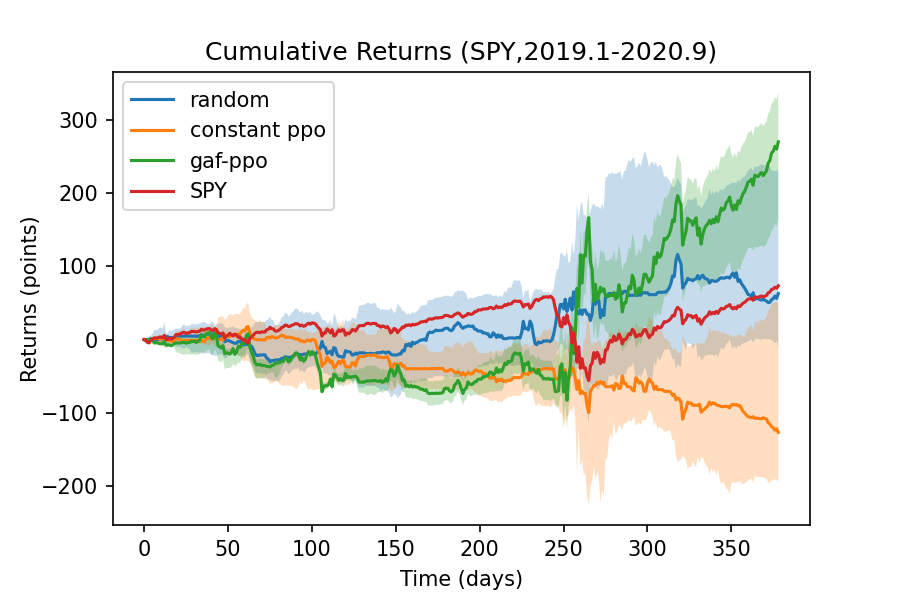}
\includegraphics[scale=0.39]{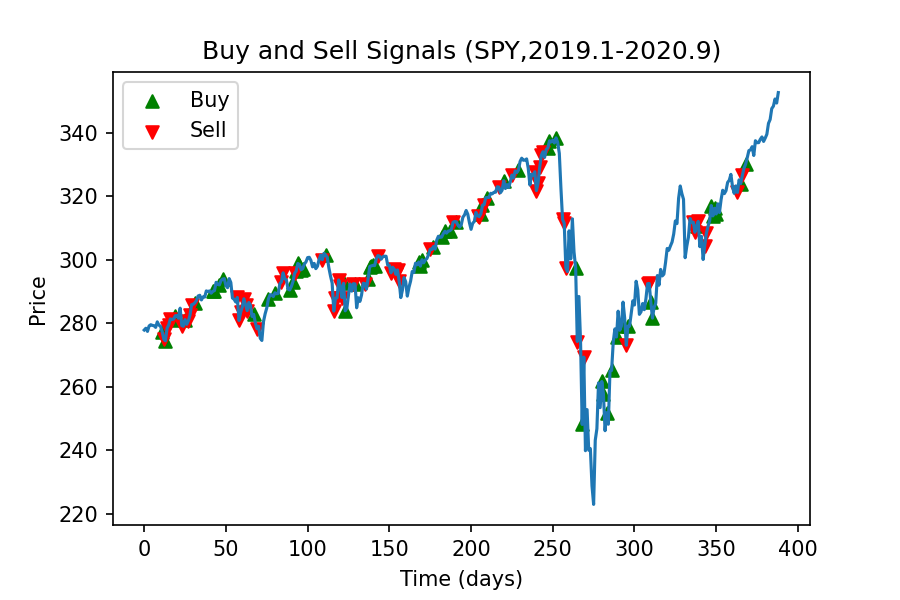}
\caption{GAF-PPO model is applied to the bull market when COVID-19 outbreak.}
\label{GAF-RL_results_UP}
\end{figure}

The data object of this test is included across time from January 2019 to September 2020. In addition, the outbreak of the COVID-19 pandemic contains during this period to test whether the stability of this trading strategy is enough to act as an effective hedge tool. In the experiment, four types of commodities in different markets, bull, bear, and stable markets selected for comparison. Since many trading platforms do not require fees, this does not consider the transaction fees. When the market volatility is more violent, the return will correlate with growth in terms of profit performance. Conversely, when the market volatility gets less, the return on profit will reduce.

Figure~\ref{GAF-RL_results_cryptocurrency} shows that the attributes of the cryptocurrency market are relatively volatile and easily affected by the news. Therefore the strategy runs for the cryptocurrency market from August to 2020 June 2021. Several times, the market rapidly fluctuates due to celebrities who support and oppose cryptocurrency. Nevertheless, the strategy can still retain sure profit and hedging ability in such a market state. The result also proves that this trading strategy has a significant response-ability in the market with severe fluctuations.

The following period is the agent's performance evaluation period. The agent uses a trading strategy to form decisions. The system accumulates the rewards obtained during the evaluation period, which served as a reference for performance. Figure~\ref{GAF-RL_results_UP} shows that the two large bull markets (Nasdaq: GOOG and NYSE: SPY) have had tremendous growth momentum over the past decade. However, they also suffered a certain degree of decline risk after the epidemic outbreak, as seen from the experimental results. During the epidemic, the strategy of the whole model will shift to the conservative and short strategy. The density of trades will decrease significantly, but it can also be excellent to catch into the market to buy in the rapid price pullback. In terms of transaction frequency, there are about 2 or 3 transactions a week. 

\begin{figure}[ht]
\centering
\includegraphics[scale=0.39]{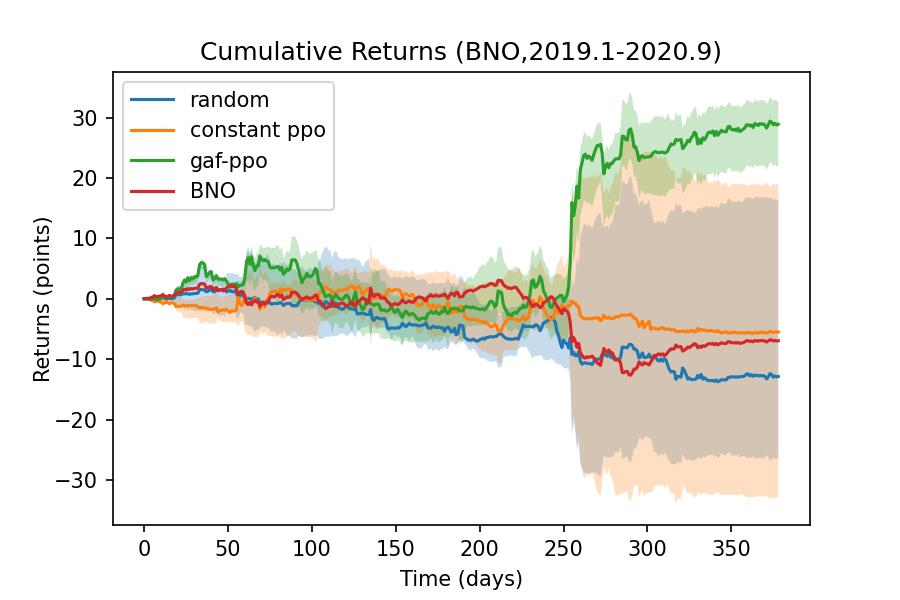}
\includegraphics[scale=0.39]{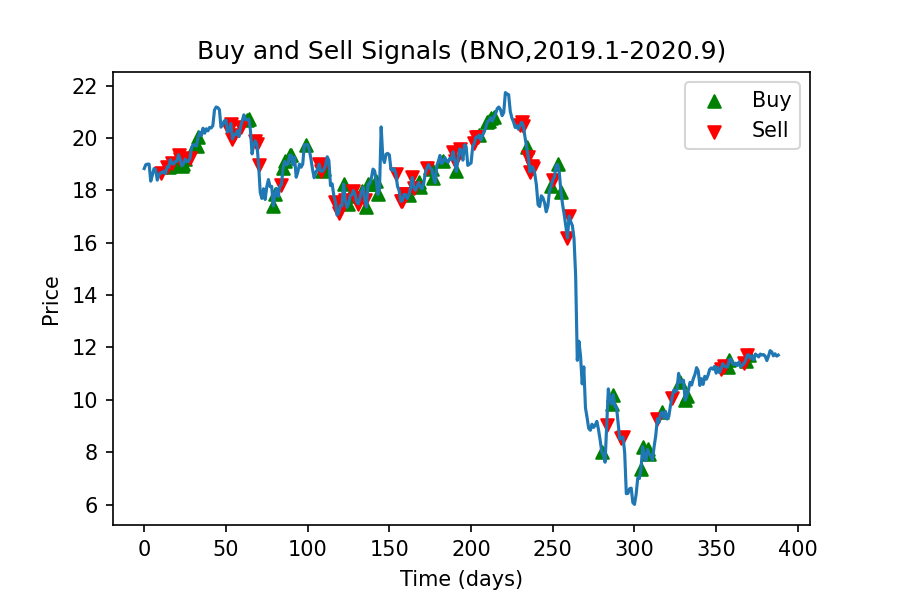}
\includegraphics[scale=0.39]{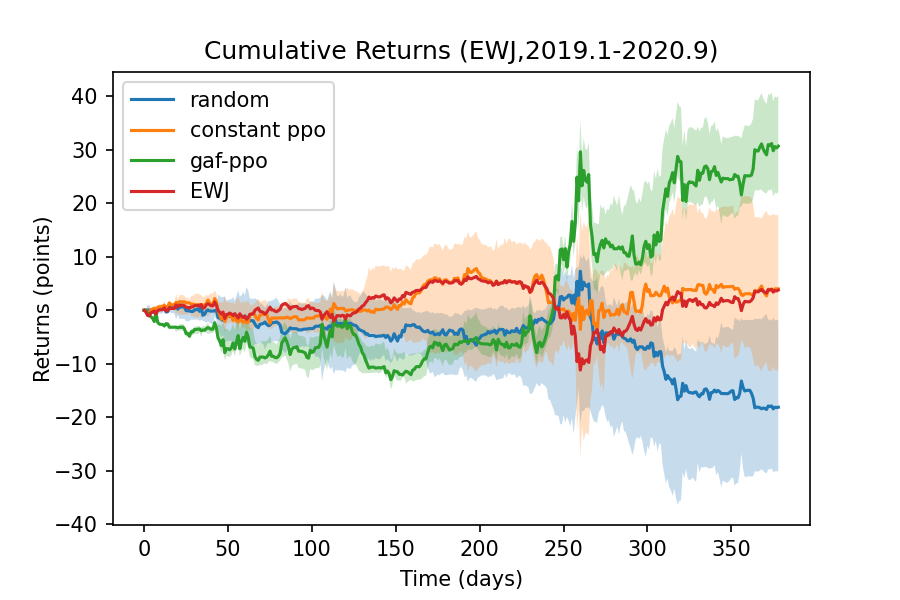}
\includegraphics[scale=0.39]{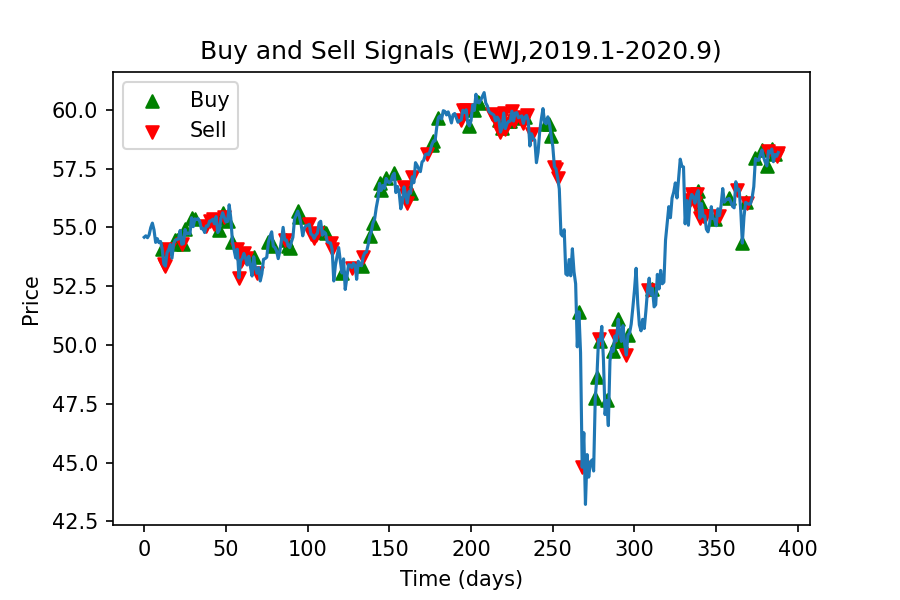}
\caption{GAF-PPO model is applied to the bear and rebound market when COVID-19 outbreak.}
\label{GAF-RL_results_DOWN}
\end{figure}

Figure~\ref{GAF-RL_results_DOWN} shows the other two assets. It has continued to enter the bear and stable market (USCF: BNO and iShares MSCI Japan ETF: EWJ). The BNO has experienced a considerable decline during this period. The PPO agent directly turned to the short direction, obtained a significant return, and bought back reasonably. The profit from this short bet accounted for most of the total gain.

However, in the part of EWJ, the initial volatility of the index did not get a significant return, representing the inadequacy of the trading strategy for the market with severe fluctuations. Nevertheless, in a word, this trading strategy based on pattern recognition can indeed accurately provide appropriate strategies as hedging operations for investors when downside risks suddenly appear.

\section{\label{sec:Discussion}Discussion}
Suppose license plate recognition is a mature commercial product in Computer Vision. In that case, our Financial Vision is the Trade Surveillance System in Financial Markets.
We want to use the concept that the camera on the road has been monitoring whether there are violations to illustrate the core contribution and use of financial vision in transactions. First, let us describe using computer vision technology to monitor violations at intersections. Such as characteristics of speeding, illegal turns, and running red lights. When a camera with license plate recognition capabilities captures the particular patterns, the traffic team will receive a notification and compare the red list with sending the photo to the offender. Using financial vision technology to monitor the price changes of the subject matter is like watching intersections. Investors set specific candlesticks patterns (for example, W bottom, M head, N-type) that require special attention and will receive Pattern Hunter's Notification. Then, how to conduct transactions is another matter. Moreover, how is this monitoring logic different from traditional price prediction?

Traditional price prediction uses the price of the past period $(0 ~ t-1)$ as the input $X$ of Supervised Learning. The price at time $t$ is the output $Y$ (or directly use the rise or rise fall as the learning target). These all imply a fatal assumption: the forecast time point in the real market in the future, the distribution of these data, and The past historical data are consistent. But in any case, it divides into a training set and test set (even if the concept of moving window uses).

When discussing intersection surveillance, the camera only cares about denying the offending vehicle's license plate. The camera does not predict whether a car will violate the regulations the next time. Therefore, when building the monitoring model, it will recognize the license plate independently. Outside of these road conditions. The road condition of this camera at the time $(0 ~ t-1)$ does not regard as input $X$ of Supervised Learning, and the road condition at time $t$ is regarded as output $Y$. Therefore, the transaction monitoring system constructs a financial vision dedicated to learning the candlesticks pattern map. When the candlesticks pattern appears (as if a red light is running, the camera will take a picture and notify the traffic team), list it, and inform the user.

We live in an era when cryptocurrency is at a crossroads of transitional adaptation that requires more sensitive approaches to adapt to the evolving financial market. By conducting Transfer Learning of cryptocurrency trading behavior, the study seeks any possibilities of implementing the trading pattern in the transactions of traditional commodities. With the Candlestick analysis, the model attempts to discover feasible rules for establishing a specific pattern. Subsequently, the trading pattern utilizes in broadening possible scenarios by Deep Reinforcement Learning. We endeavor to comprehend profitability strategies and market sentiment in various conditions to satisfy voluminous risk aversion demands in the market. The experiment results reveal that more accurate judgments can make when the commodity price fluctuates volatilely.
On the contrary, human beings tend to be influenced by emotional bias on most occasions. The inclination acquaints human beings with more stable and long-term markets. Human beings are adapted to be blindly spotted and miscalculate the best entry point when volatile movement occurs. Integrating the application of the machine learning model can effectively compensate for human deficiency and enhance the efficiency of the trading process. Nevertheless, optimizing the trading frequency to consolidate the liquidity of different financial assets is another issue that deserves more attention in the dynamic financial market.

\section{\label{sec:Conclusion}Conclusion}
For the current wise investment, investors want to predict the future transaction price or ups and downs directly. The fatal assumption is that the training data set is consistent with the data distribution that has not occurred in the future. However, the natural world will not let us know whether the subsequent data distribution will change.

Because of this, even if researchers add a moving window to the training process, it is inevitable that "machine learning obstacles-prediction delay" will occur. Just search for "machine learning predict stock price," and researchers can find articles full of pits, all of which have this shortcoming. Therefore, our first contribution is not to make future predictions but to focus on the current "candlesticks pattern detection," such as Engulfing Pattern, Morning Star,\ldots.

However, the "candlesticks pattern" is usually a sensational description. It cannot become a stylized trading strategy if investors cannot write a program to enumerate all the characteristics. Even if a trader has a sense of the market and knows which patterns have to enter and exit, he cannot keep his eyes on all the investment targets. Moreover, our second contribution focuses on detecting trading entry and exit signals combined with related investment strategies.

Finally, we found from experiments that the 15-minute price data of Ethereum train through transfer learning is suitable for US stock trading. Compared to the top ten most popular ETFs, the experimental results demonstrate superior performance. This study focuses on financial vision, Explainable methods, and links to their programming implementations. We hope that our paper will reference superhuman performances and why the decisions are in the trading system.

\section*{Appendix}
The eight patterns used in this study will describe entirely in this section.
The following eight figures illustrate the critical rules each pattern requires. The white candlestick represents a rising price on the left-hand side of each figure, and the black candlestick represents a dropping price. The arrow indicates the trend. The upward arrow indicates a positive direction, and the downward arrow indicates a negative trend. The text descriptions on the right-hand side are the fundamental rules referred to from The Major Candlestick Signals.~\cite{MajorSignals}.

\begin{figure*}[ht]
  \centering
    \includegraphics[width=0.45\textwidth]{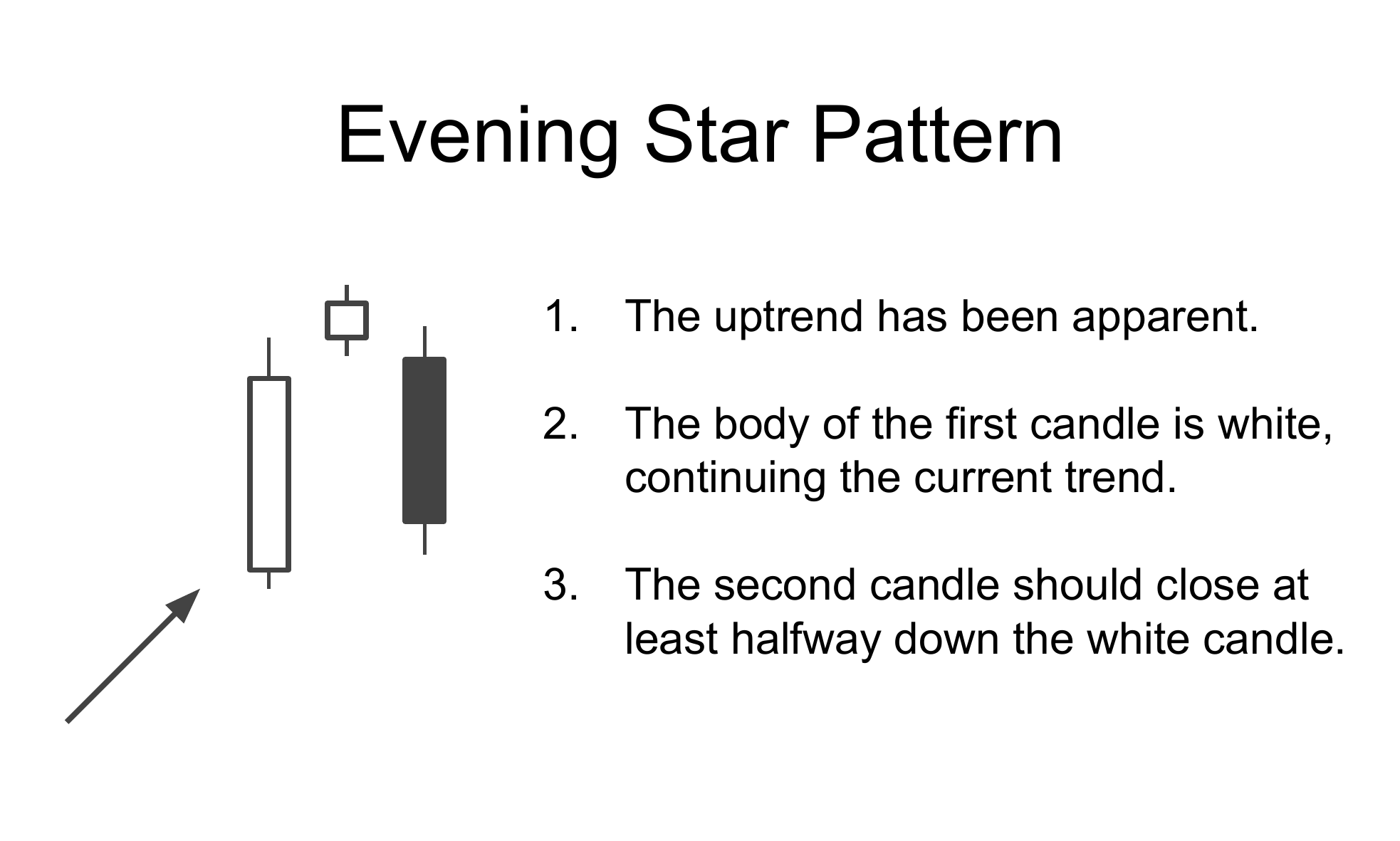}
    \includegraphics[width=0.45\textwidth]{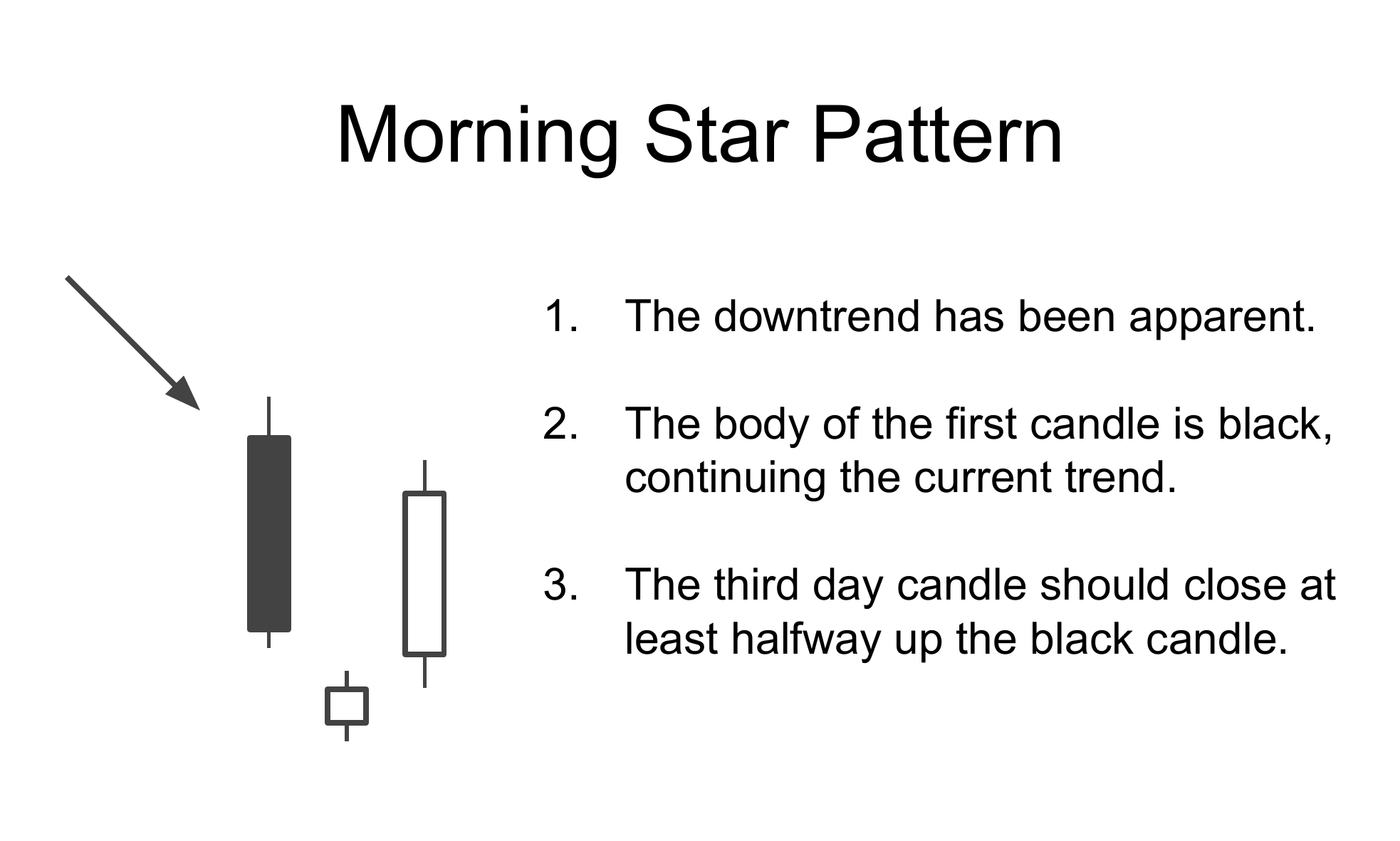}
    \includegraphics[width=0.45\textwidth]{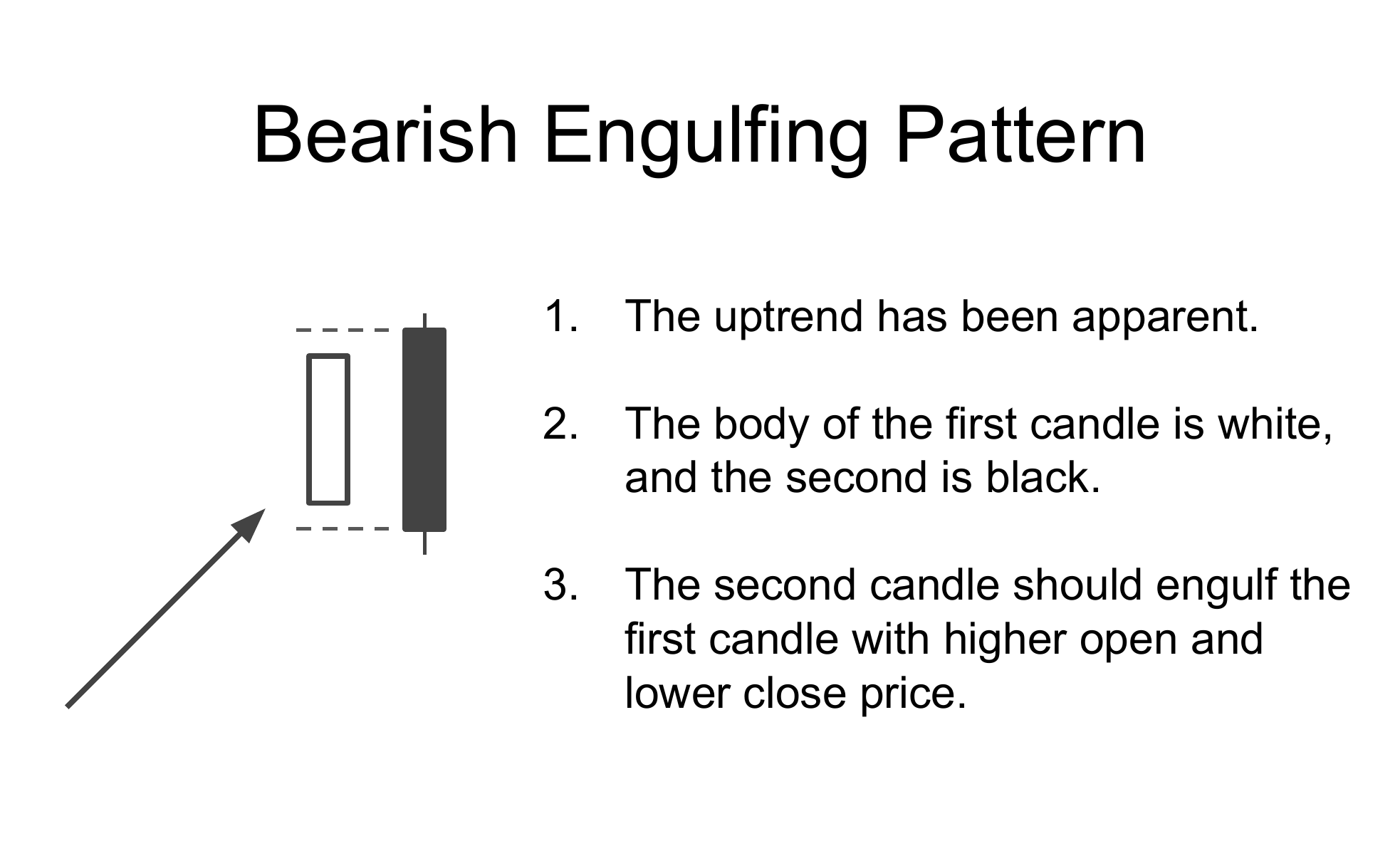}
    \includegraphics[width=0.45\textwidth]{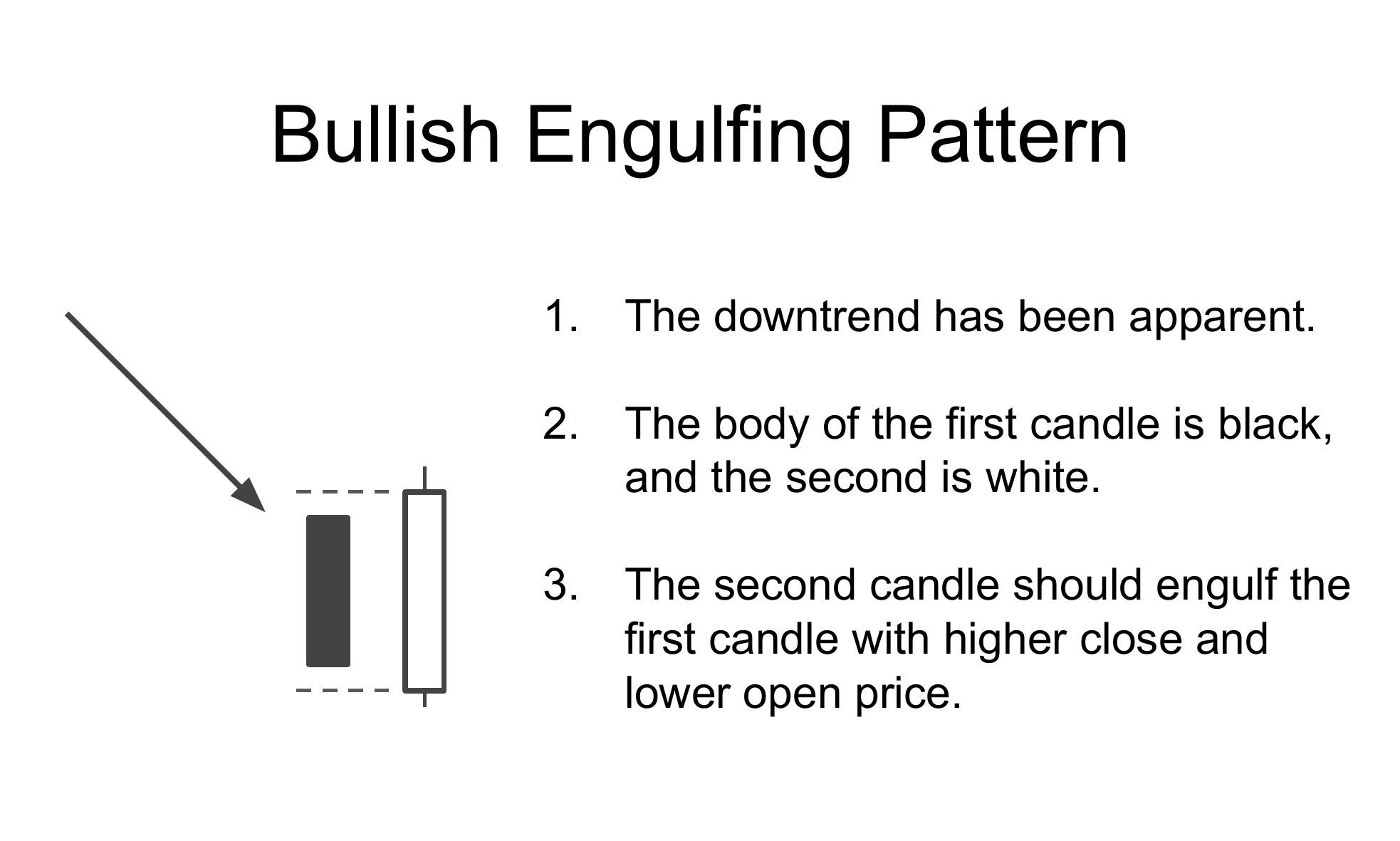}
\end{figure*}

\begin{figure*}[ht]
  \centering
    \includegraphics[width=0.45\textwidth]{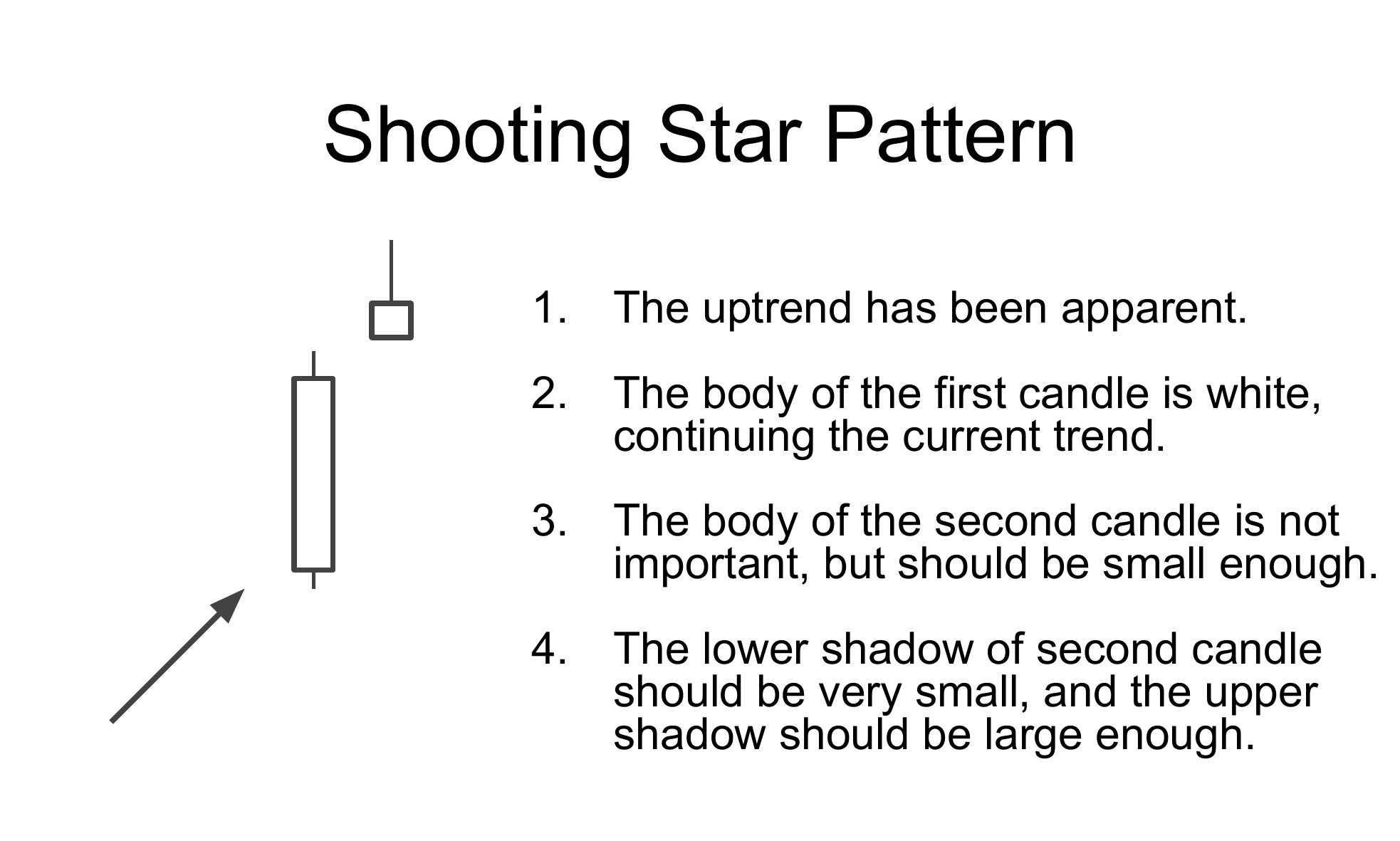}
    \includegraphics[width=0.45\textwidth]{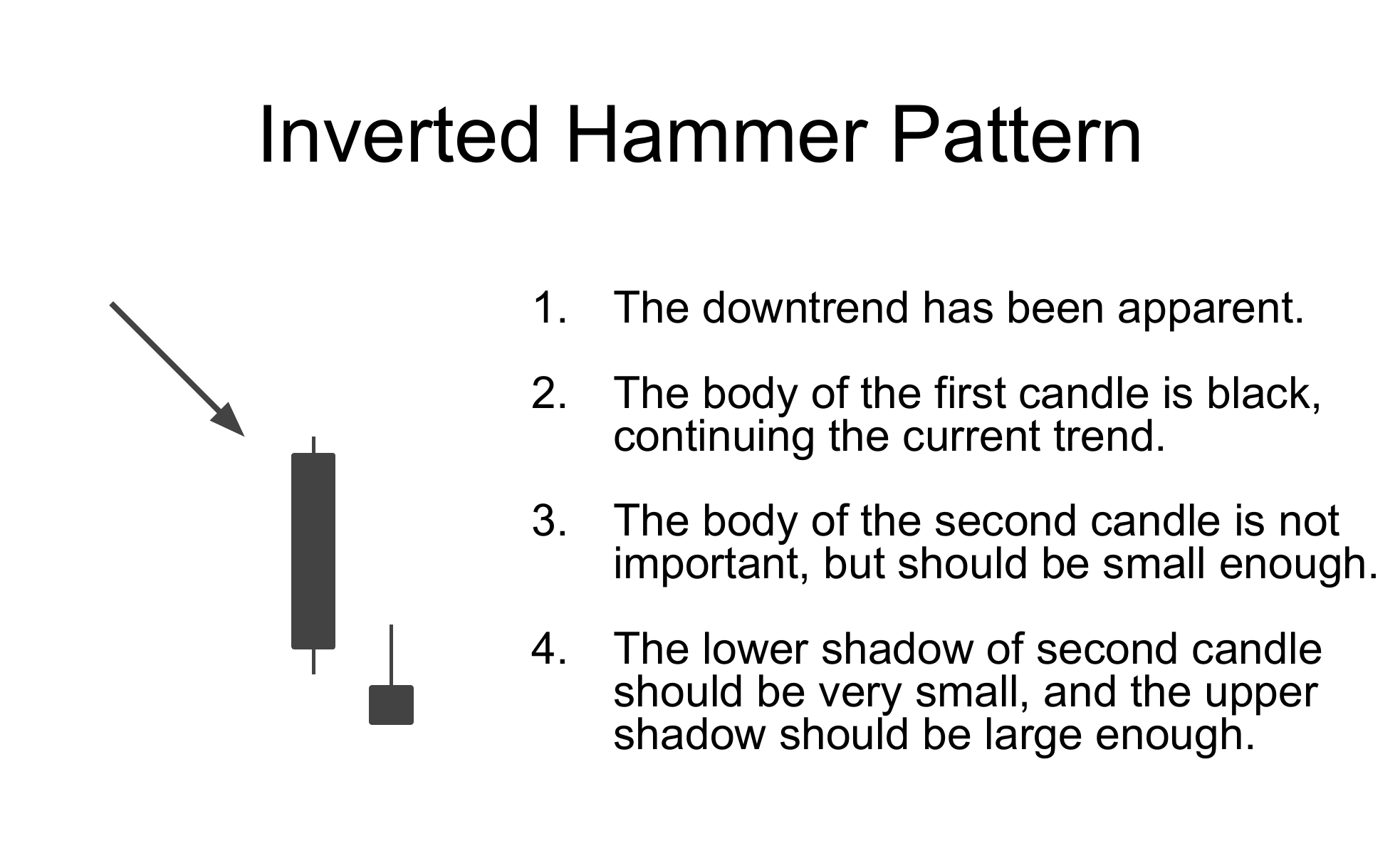}
    \includegraphics[width=0.45\textwidth]{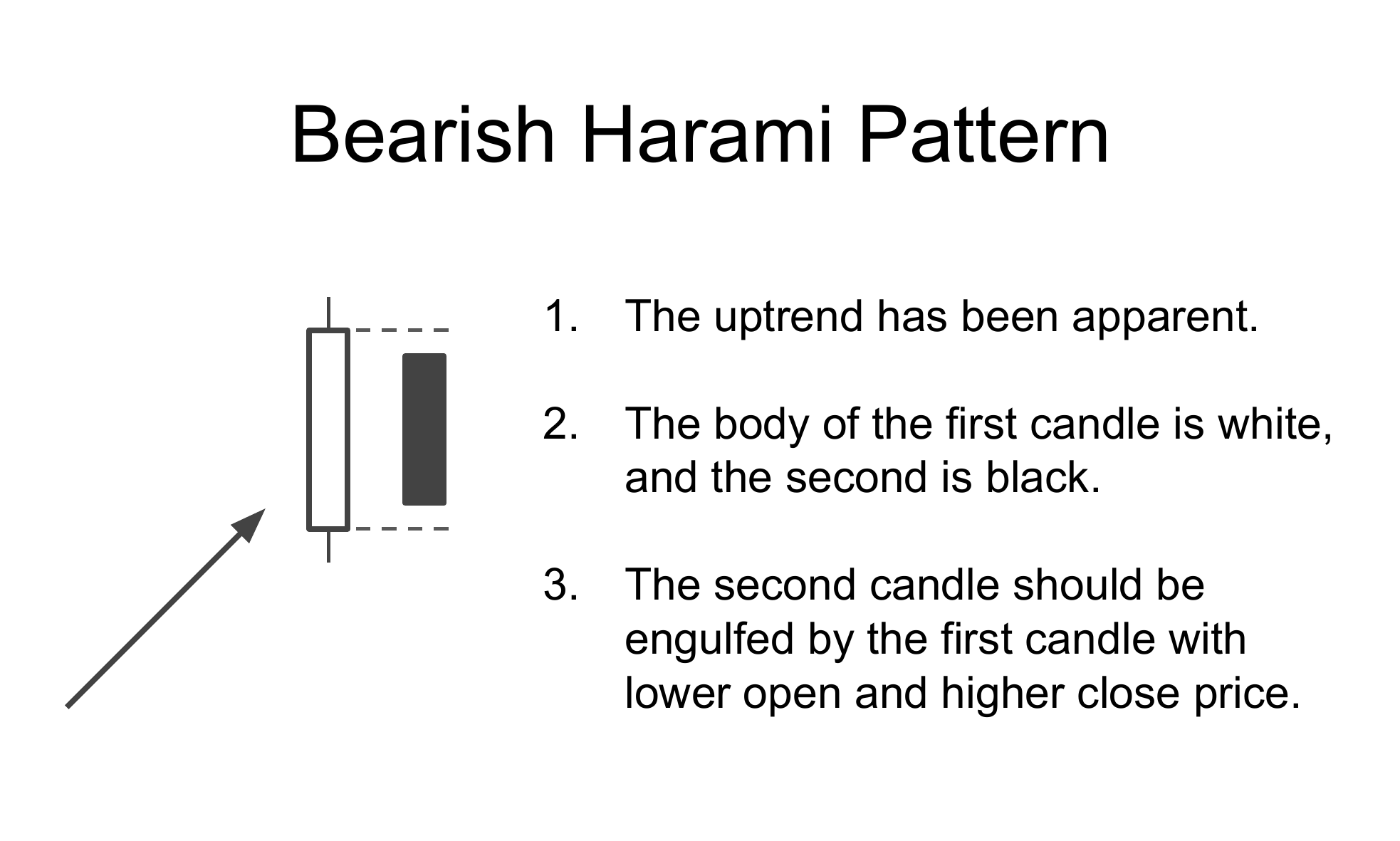}
    \includegraphics[width=0.45\textwidth]{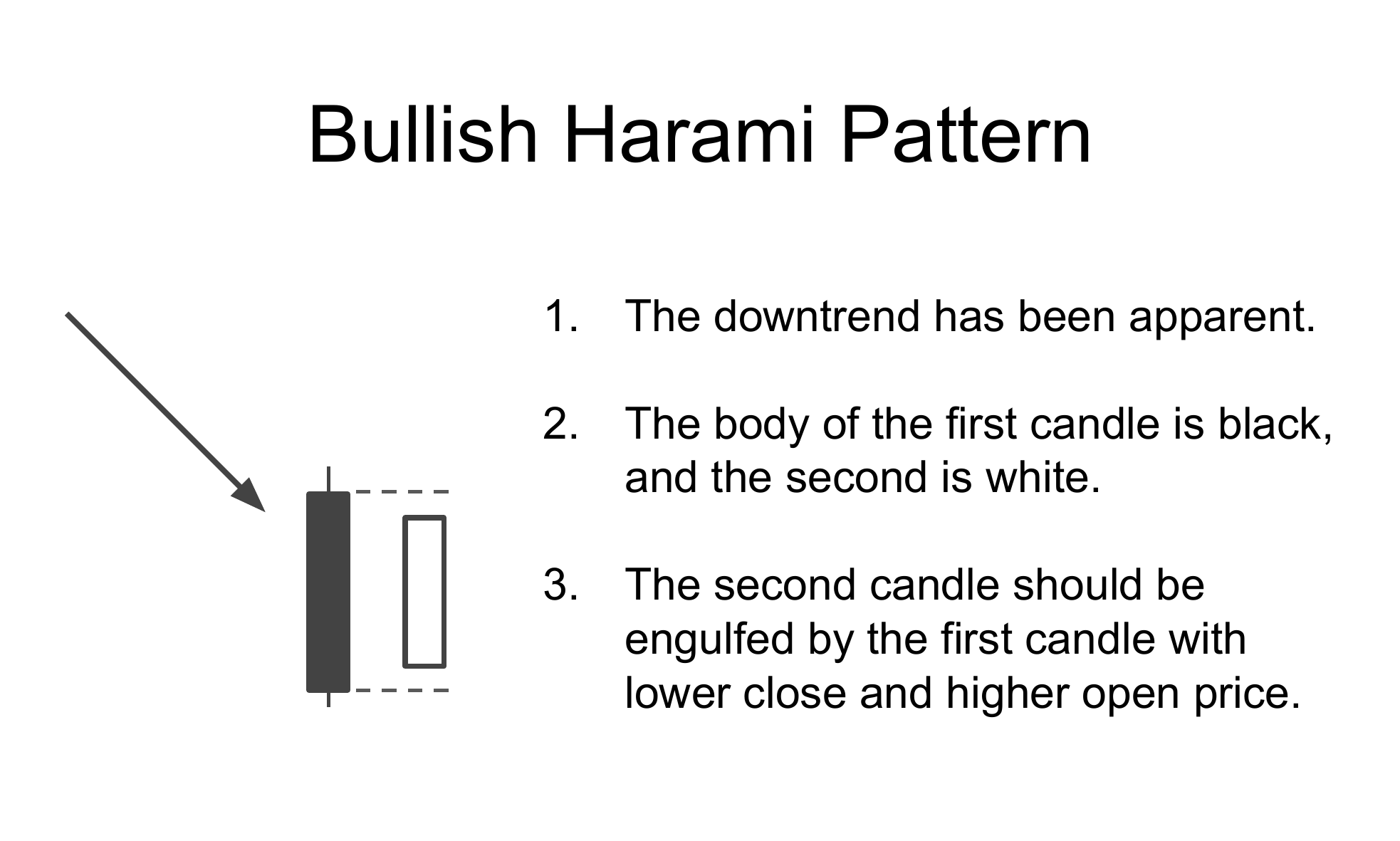}
\end{figure*}

\bibliographystyle{bmc-mathphys}
\bibliography{template}
\end{document}